\documentclass[10pt,twocolumn,letterpaper]{article}

\usepackage{iccv}
\usepackage{times}
\usepackage{epsfig}
\usepackage{graphicx}
\usepackage{amsmath}
\usepackage{amssymb}

\usepackage[numbers,sort]{natbib}
\usepackage{color}
\usepackage{booktabs}
\usepackage[pagebackref=true,breaklinks=true,letterpaper=true,colorlinks,linkcolor=blue,citecolor=blue,bookmarks=false]{hyperref}
\usepackage{cleveref}
\usepackage[accsupp]{axessibility}  %

\iccvfinalcopy %

\def\iccvPaperID{1281} %

\newcommand{\first}[1]{\textbf{#1}}
\newcommand{\second}[1]{\underline{#1}}

\begin{document}
\title{3D Human Texture Estimation from a Single Image with Transformers}
\author{Xiangyu Xu
	\qquad
	Chen Change Loy\\
	S-Lab, Nanyang Technological University \\
{\tt\small xiangyu.xu@ntu.edu.sg, ccloy@ntu.edu.sg}
}

\maketitle
\begin{abstract}
	We propose a Transformer-based framework for 3D human texture estimation from a single image.
	The proposed Transformer is able to effectively exploit the global information of the input image, overcoming the limitations of existing methods that are solely based on convolutional neural networks. 
	In addition, we also propose a mask-fusion strategy to combine the advantages of the RGB-based and texture-flow-based models.
	We further introduce a part-style loss to help reconstruct high-fidelity colors without introducing unpleasant artifacts.
	Extensive experiments demonstrate the effectiveness of the proposed method against state-of-the-art 3D human texture estimation approaches both quantitatively and qualitatively.
	The project page is at \url{https://www.mmlab-ntu.com/project/texformer}.
\end{abstract}

\section{Introduction} \label{sec: introduction}
\begin{figure}[t]
	\footnotesize
	\begin{center}
		\begin{tabular}{c}
						\hspace{-2mm}
			\includegraphics[width = 0.96\linewidth]{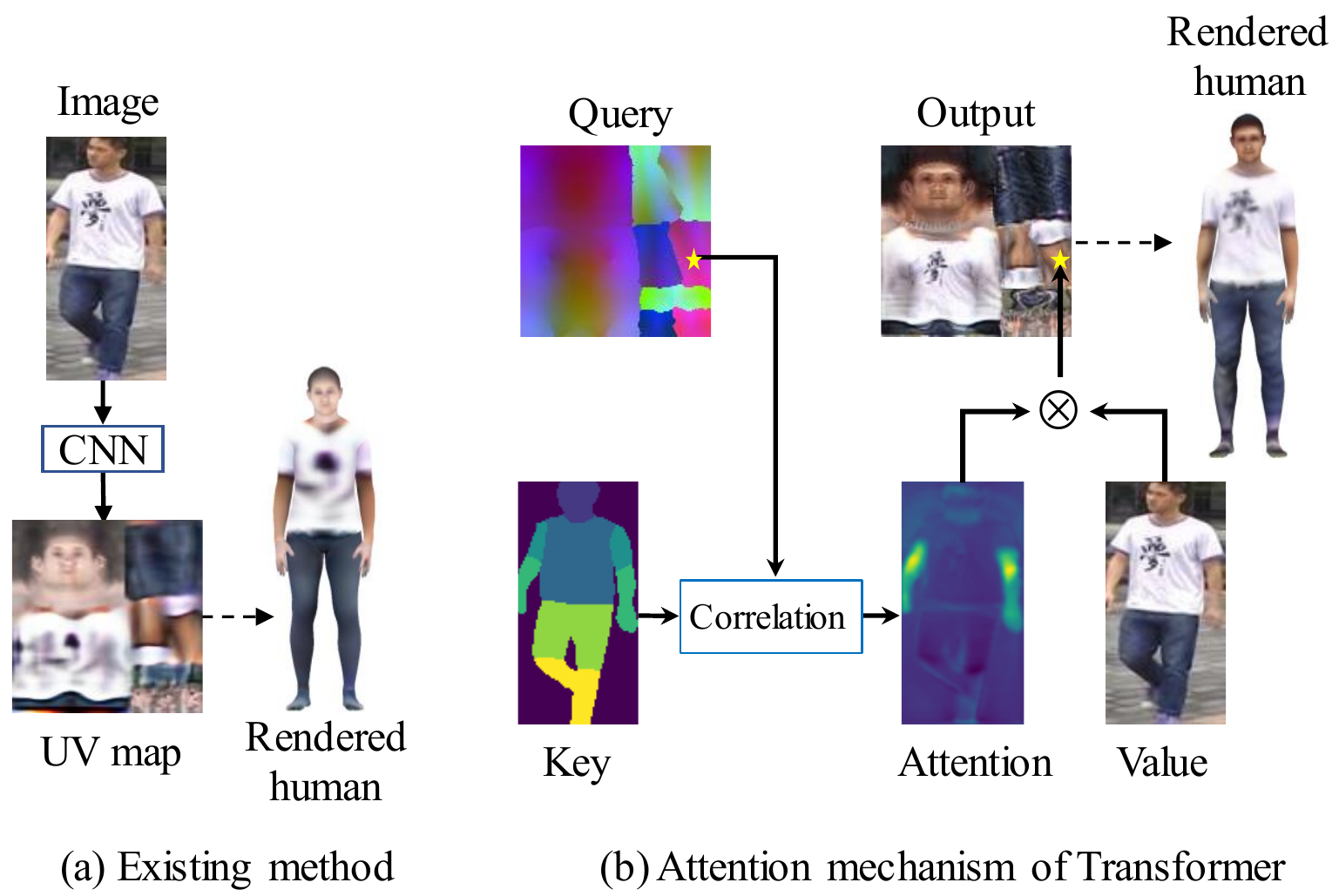} 
		\end{tabular}
	\end{center}
	\vspace{-3mm}
	\caption{We propose a Transformer for 3D human texture estimation from a single image. Compared with the existing method~\cite{wang2019re} that solely relies on CNNs (a), the proposed Transformer uses the attention mechanism to more effectively exploit global information (b), which leads to higher-quality 3D human texture estimation. See the text for more detailed explanations.
	}
	\vspace{-6mm}
	\label{fig: teaser}
\end{figure}

In this paper, we study the problem of estimating 3D human texture from a monocular image. 
This is an important problem that plays a key role in single-image 3D human reconstruction and has wide applications in virtual and augmented reality, film industry, gaming, and biometrics.
Most of existing methods for this problem~\cite{wang2019re,xu2021_3d,zhao2020human,kanazawa2018learning,li2020self,goel2020shape} use deep convolution neural networks (CNNs) to predict 3D human texture (\ie, a UV map) from the input image (Figure~\ref{fig: teaser}(a)).
While these methods have achieved impressive results, their network architectures suffer from an inherent shortcoming: convolution layers are by design local operations and inefficient in processing global information that is crucial in 3D human texture estimation.
More specifically, the input and output in this task do not have strictly-aligned spatial correspondences and may even have totally different shapes as shown in Figure~\ref{fig: teaser}(a). 
This is in sharp contrast to 2D computer vision tasks such as image super-resolution~\cite{xxy-iccv17} and image-to-image translation~\cite{isola2017image} where the input and output are well aligned.

We believe that one should not simply follow the common practice of using only CNNs for the 3D task in this work, and global operations should be introduced for better 3D human texture reconstruction.
Note that some recent approaches~\cite{xu2021_3d,zhao2020human} attempt to address this issue with fully-connected layers (or MLPs), which however often leads to loss of fine spatial information. 
Moreover, this information loss cannot be easily remedied by skip connections~\cite{ronneberger2015u} due to the misalignment of the different-layer features.
A successful solution should be able to more effectively distribute features of the input into suitable locations in the UV space while preserving fine spatial information.

Towards this end, we propose a Transformer-based framework for 3D human texture estimation from a single image. 
The Transformer allows processing information of the input in a global manner, which is in particular suitable for our task.
At the core of the Transformer is an attention module that involves three basic components: Query, Key, and Value.
For the Query, we use a pre-computed color map that has the same shape as the output UV map.
Each pixel in the query map corresponds to a vertex on 3D human mesh~\cite{loper2015smpl}.
The Value is the input image that has all the source pixels.
For the Key, we use a 2D part-segmentation map obtained with an off-the-shelf model~\cite{huang2018eanet}, which implies the mapping from the image to the UV space~\cite{zhao2020human,mir2020learning}.

For an intuitive understanding of the relationship between these three components, we elucidate the working mechanism of our Transformer in Figure~\ref{fig: teaser}(b).
A pixel in the Query (marked as yellow star) is first used to correlate with the Key, which produces an attention map. 
With this attention map, the source information in the Value can then be effectively aggregated by weighted averaging to generate the corresponding pixel in the output UV map.
Such an attention mechanism allows us to exploit global information of the input image without losing fine details, which is the key factor that distinguishes our method from existing algorithms.
Note that the above explanations have been simplified for ease of understanding. 
As will be introduced in Section~\ref{sec: algorithm}, we use more channels for the Value and Key in real implementation and perform multi-head attention~\cite{vaswani2017attention} in feature space.

To summarize, we make the following contributions:

\noindent 
1) We propose a Transformer-based framework, termed as {Texformer}, for 3D human texture estimation from a single image. Based on the attention mechanism, the proposed network is able to effectively exploit global information of the input. It naturally overcomes the limitations of existing algorithms that solely rely on CNNs and effectively facilitates higher-quality 3D human texture reconstruction.

\noindent
2) Existing algorithms output either RGB values~\cite{wang2019re,xu2021_3d} or texture flow~\cite{kanazawa2018learning,zhao2020human} to synthesize the final UV map. We analyze the limitations of these two strategies and propose a new method to combine the best of both worlds. We show that the proposed method is able to significantly reduce visual artifacts while preserving fine details.

\noindent
3) The estimated textures of previous approaches often suffer from noticeable color differences from the input. To remedy this issue, we propose a part-style loss that enforces the Gram-matrix similarity~\cite{gatys2016image} for each human body part and encourages closer appearances to the input image.

Extensive experiments on the Market-1501 dataset~\cite{zheng2015scalable} demonstrate the effectiveness of the proposed method against state-of-the-art 3D human texture estimation approaches both quantitatively and qualitatively.

\section{Related Work}
\noindent \textbf{3D human texture estimation.}
Recent years have witnessed significant progress in the field of 3D human texture estimation~\cite{huang2020arch,lazova2019360,saito2019pifu,natsume2019siclope,zheng2020pamir,neverova2018dense,alldieck2018detailed,bhatnagar2019multi,mir2020learning,wang2019re,xu2021_3d,zhao2020human,kanazawa2018learning,li2020self,goel2020shape,alldieck2018video,alldieck2019learning,li2020robust,zhi2020texmesh}.
Some methods~\cite{alldieck2018detailed,alldieck2018video,bhatnagar2019multi,alldieck2019learning,mir2020learning} solve this problem by taking multi-view images as input, and the 3D human texture can be synthesized by merging textures from different views with Graph-cut~\cite{boykov2001fast}, median filters~\cite{alldieck2018video}, or deep learning~\cite{mir2020learning}.
However, these methods cannot be easily used for single-image 3D human reconstruction~\cite{zhu2020reconstructing,pumarola20193dpeople,saito2020pifuhd,zheng2019deephuman,alldieck2019tex2shape}.
More recent approaches in this direction use RGBD videos~\cite{li2020robust,zhi2020texmesh}, which poses even higher demand for the input.

For applications in single-image 3D human reconstruction~\cite{zhu2020reconstructing,pumarola20193dpeople,saito2020pifuhd,zheng2019deephuman,alldieck2019tex2shape}, 
another line of approaches use deep neural networks to reconstruct 3D human textures from a monocular image~\cite{huang2020arch,lazova2019360,saito2019pifu,natsume2019siclope,zheng2020pamir}.
However, these methods usually need 3D supervision obtained via 3D scanning, which is both time-consuming and labor-intensive.
Moreover, some methods~\cite{lazova2019360,neverova2018dense} require high-quality dense human pose estimation for extracting partially observed textures, which would be challenging for in-the-wild images.

More closely related to this work, some recent methods aim to address the problem of single-image texture reconstruction without 3D labels~\cite{wang2019re,xu2021_3d,zhao2020human,kanazawa2018learning,li2020self,goel2020shape}, which are generally more applicable for real-world usage.
For instance, Wang~\etal~\cite{wang2019re} propose a self-supervised training pipeline by building upon the success of existing 3D human mesh reconstruction algorithms~\cite{kolotouros2019spin,kocabas2019vibe,kanazawa2018end,kanazawa2019learning,huang2018deep,xu20203d}.
Kanazawa~\etal~\cite{kanazawa2018learning} also learn the 3D textures in a self-supervised manner and use texture flow to directly sample pixels from the input image.
In addition, Zhao~\etal~\cite{zhao2020human} exploit human part segmentation as input and use multi-view data as extra training labels for better performance.
While these methods have achieved promising progress in this task, they usually rely on CNNs that cannot well exploit global information of the input. 
In contrast, we propose a Transformer-based framework that effectively overcomes this issue.
In addition, we introduce a mask-fusion strategy to combine the advantages of both RGB predictions and texture flow.
We also propose part-style loss and face-structure loss to further improve the texture reconstruction results.

\noindent \textbf{Transformers.}
Thanks to the remarkable ability to handle long-range information, the Transformer has become the dominant architecture in natural language processing~\cite{vaswani2017attention,devlin2018bert}.
Recently, it has also been introduced in computer vision, and the applications include object detection~\cite{carion2020end}, image classification~\cite{dosovitskiy2020image}, image restoration~\cite{chen2020pre}, and hand pose estimation~\cite{huang2020hand}. 
We refer to \cite{han2020survey} for a more comprehensive review of vision transformers.

In this work, we demonstrate that the attention mechanism makes the Transformer particularly suitable for the task of 3D human texture estimation.
We provide new insights in designing the Transformer for this task, and the proposed network achieves better performance than the state-of-the-art approaches.
We also show that a common Transformer structure without specific design cannot perform as well as the proposed architecture.

\begin{figure*}[t]
	\footnotesize
	\begin{center}
		\begin{tabular}{c}
			\includegraphics[width = 0.9\linewidth]{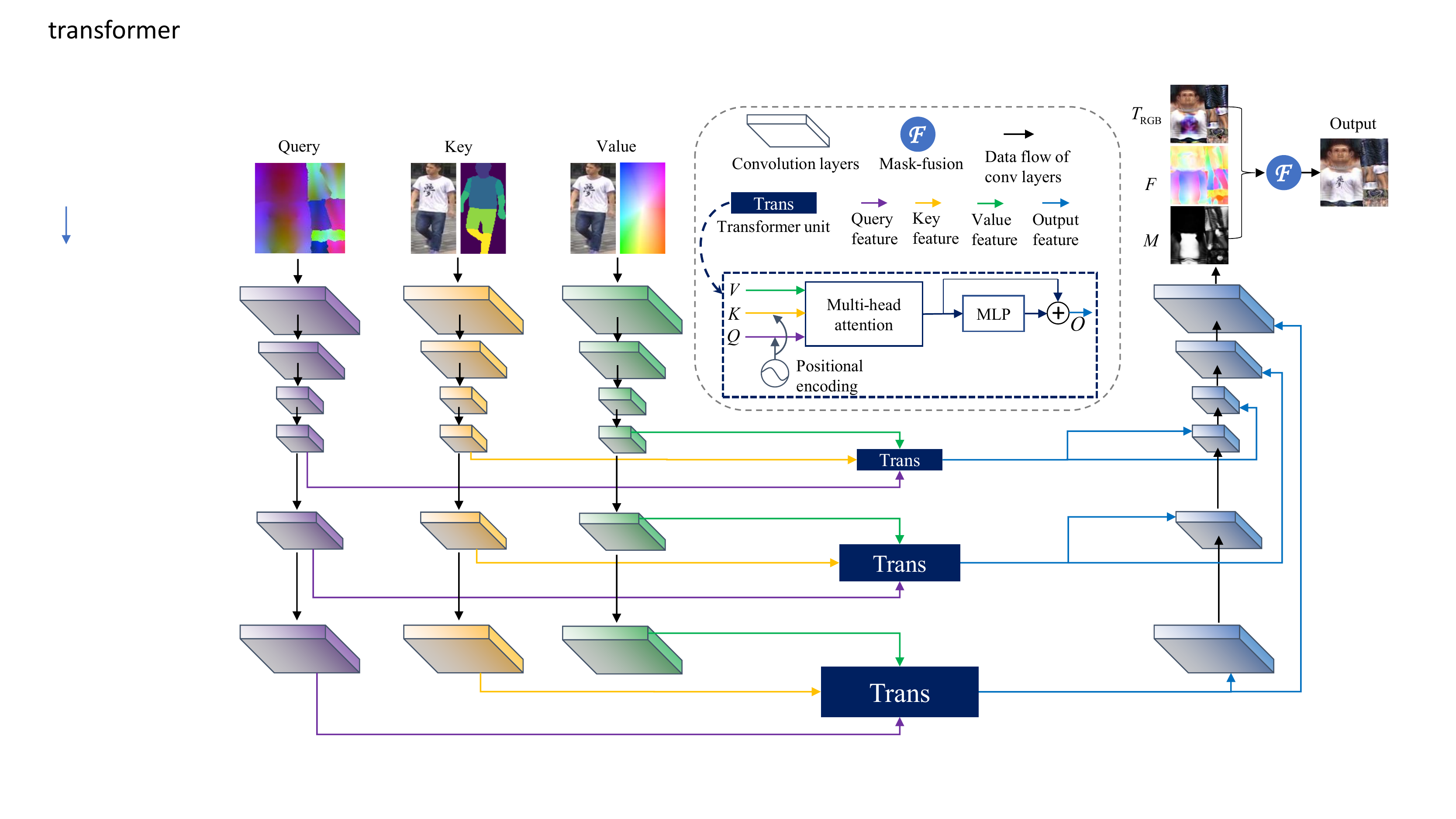} 
		\end{tabular}
	\end{center}
	\vspace{-3mm}
	\caption{Overview of the proposed Texformer. The Query is a pre-computed color encoding of the UV space obtained by mapping the 3D coordinates of a standard human body mesh to the UV space. The Key is a concatenation of the input image and the 2D part-segmentation map. The Value is a concatenation of the input image and its 2D coordinates.
		We first feed the Query, Key, and Value into three CNNs to transform them into feature space.
		Then the multi-scale features are sent to the Transformer units to generate the Output features.
		The multi-scale Output features are processed and fused in another CNN, which produces the RGB UV map $T_\text{RGB}$, texture flow $F$, and fusion mask $M$.
		The final UV map is generated by combining $T_\text{RGB}$ and the textures sampled with $F$ using the fusion mask $M$.
		Note that we have skip connections between the same-resolution layers of the CNNs similar to \cite{xu2020learning_stpan} which have been omitted in the figure for brevity.
	}
	\vspace{-5mm}
	\label{fig: overview}
\end{figure*}

\section{Methodology} \label{sec: algorithm}
We propose a Transformer-based framework, termed as Texformer, for 3D human texture estimation from a single image.
An overview of the Texformer is shown in Figure~\ref{fig: overview}.
A brief explanation of the framework is provided in the figure caption. Next, we provide more detailed explanations of the Texformer as well as our loss functions.

\subsection{Texformer} \label{sec: transformer}
\noindent \textbf{Query.} As introduced in Section~\ref{sec: introduction}, the Query of our model represents a color-encoding of the output UV space, and each pixel in the Query map should characterize a vertex on the 3D human mesh.
To this end, we use the 3D coordinates of a standard human body mesh from the SMPL model~\cite{loper2015smpl} as the color encoding of each vertex. The color values of each pixel in the Query can be obtained by first mapping the 3D coordinates to the UV space and then interpolating the coordinates with KD-Tree~\cite{maneewongvatana1999analysis}. %
We pre-compute the Query map before network training, and the same color encoding of the UV space is used for all input images.

\noindent \textbf{Key.} Recall that the Key is used to correlate with the Query elements to obtain the attention map for the input (Figure~\ref{fig: teaser}), which is crucial for connecting the image space and the UV space.
Motivated by recent studies~\cite{zhao2020human,mir2020learning}, we use 2D part-segmentation as the Key to learn a mapping from the input image to the output UV map.
Our 2D part-segmentation is obtained with an off-the-shelf model~\cite{huang2018eanet}. 
As shown in Figure~\ref{fig: overview}, we also include the input image as additional channels of the Key to provide contextual information.

\noindent \textbf{Value.} The Value represents the source information indexed by the Key and is aggregated into the UV space to generate the output using the attention map. 
There are two possible options for the Value depending on the form of the Transformer output: 1) using the RGB input image as the Value when the Transformer directly outputs the RGB UV map; 2) using the flow field of the image, \ie, 2D coordinates for each pixel, when the model first produces the texture flow and then generates the UV map by sampling the input image with the predicted flow.
As will be introduced later in this section, our Transformer predicts both RGB values and texture flow, and thus, we concatenate both the RGB image and the flow field as the Value of our model (Figure~\ref{fig: overview}).

The Query, Key, and Value cannot be directly used by the Transformer with their raw forms; 
this is especially the case for the Query and Key which will be compared in the attention module and thereby should have the same feature dimension.
Thus, we feed them in CNNs to transform them into feature space as shown in Figure~\ref{fig: overview}.
Then the resulting features $Q \in \mathbb{R}^{vu \times d}$, $K \in \mathbb{R}^{hw \times d}$, and $V \in \mathbb{R}^{hw \times c}$ are sent into the Transformer unit, which produces the Output features $O \in \mathbb{R}^{vu \times c}$.
Here, $v, u$ are the height and width of the output UV map, $h, w$ are the height and width of the input image, while $d$, $c$ are the feature dimensions of $K$ and $V$, respectively.
Note that the CNN of $Q$ is used to find a better representation of the fixed Query map and only needed in the training phase. During deployment, we can drop the Query CNN by pre-computing the feature encodings.

The Transformer unit in Figure~\ref{fig: overview} is the central part of the proposed network, which effectively distributes the image features into suitable locations of the UV map and enables global information interchange between the input space and the output space. 
Specifically, it can be written as:
\begin{align}\label{eq: trans unit}
	O = f_\text{res-MLP}(f_\text{Attn}(Q+E_Q, K+E_K, V)),
\end{align}
where $f_\text{Attn}$ is the multi-head attention module from~\cite{vaswani2017attention}. 
$f_\text{res-MLP}$ is a two-layer MLP with a residual connection between its input and output.
$E_Q$ and $E_K$ are sinusoidal positional encodings~\cite{vaswani2017attention} for the Query and Key features, respectively.

\noindent \textbf{Low-rank attention.}
As introduced in~\cite{vaswani2017attention}, the multi-head attention $f_\text{Attn}$ is based on a single-head attention module:
\begin{align}\label{eq: single-attn}
	f_\text{single}(\tilde{Q}, \tilde{K}, \tilde{V}) = \text{softmax}({\tilde{Q}\tilde{K}^\top}/{\alpha})\tilde{V},
\end{align}
where $\tilde{Q} \in \mathbb{R}^{\tilde{v}\tilde{u} \times \tilde{d}}$, $\tilde{K} \in \mathbb{R}^{\tilde{h}\tilde{w} \times \tilde{d}}$, and $\tilde{V} \in \mathbb{R}^{\tilde{h}\tilde{w} \times \tilde{c}}$. 
The denominator $\alpha$ is incorporated to avoid large values. 
For clarity, we do not use the symbols $Q$, $K$, and $V$ here as these features need to be first processed by linear projection layers in the multi-head attention $f_\text{Attn}$~\cite{vaswani2017attention}.

One important problem of the normal attention formulation is the high memory complexity ($\mathcal{O}(\tilde{v}\tilde{u}\tilde{h}\tilde{w})$ for Eq.~\ref{eq: single-attn}), which makes the training infeasible for common GPUs. 
To remedy this issue, we propose a low-rank attention (LoRA) module inspired by~\cite{xu2020learning}:
\begin{align}\label{eq: low-rank-attn}
f_\text{LoRA}(\tilde{Q}, \tilde{K}, \tilde{V}) = {\tilde{Q}\tilde{K}^\top\tilde{V}}/{\alpha},
\end{align}
where the softmax function is removed, and the attention is approximated by pure matrix multiplication. 
This strategy allows more efficient computations by manipulating the order of the matrix multiplications~\cite{xu2020learning} (\ie, computing $\tilde{K}^\top \tilde{V}$ first). 
The memory footprint can be significantly reduced to $\mathcal{O}(\max(\tilde{v}\tilde{u},\tilde{h}\tilde{w}) \cdot \max(\tilde{d}, \tilde{c}))$, which makes the training more feasible.
Note that \cite{xu2020learning} uses this low-rank strategy for matrix factorization while we apply it to multi-head attention for efficient Transformers.

\noindent \textbf{Multi-scale feature fusion.} 
Instead of applying the Transformer unit at one single scale, we compute the Output features with a three-level feature pyramid.
For the first two levels where the features have large spatial sizes, we use the proposed LoRA (Eq.~\ref{eq: low-rank-attn}) to reduce memory cost.
For the small features on the third level, we simply use the softmax attention module (Eq.~\ref{eq: single-attn}).
Then the multi-scale Output features $O_i, i=1,2,3$ can be easily fused within a CNN as shown in Figure~\ref{fig: overview}.
This multi-scale approach can help the proposed Transformer better exploit contextual information and predict higher-quality human textures for invisible regions (Figure~\ref{fig: ablation}).

\begin{figure}[t]
	\footnotesize
	\begin{center}
		\begin{tabular}{c}
			\includegraphics[width = 0.9\linewidth]{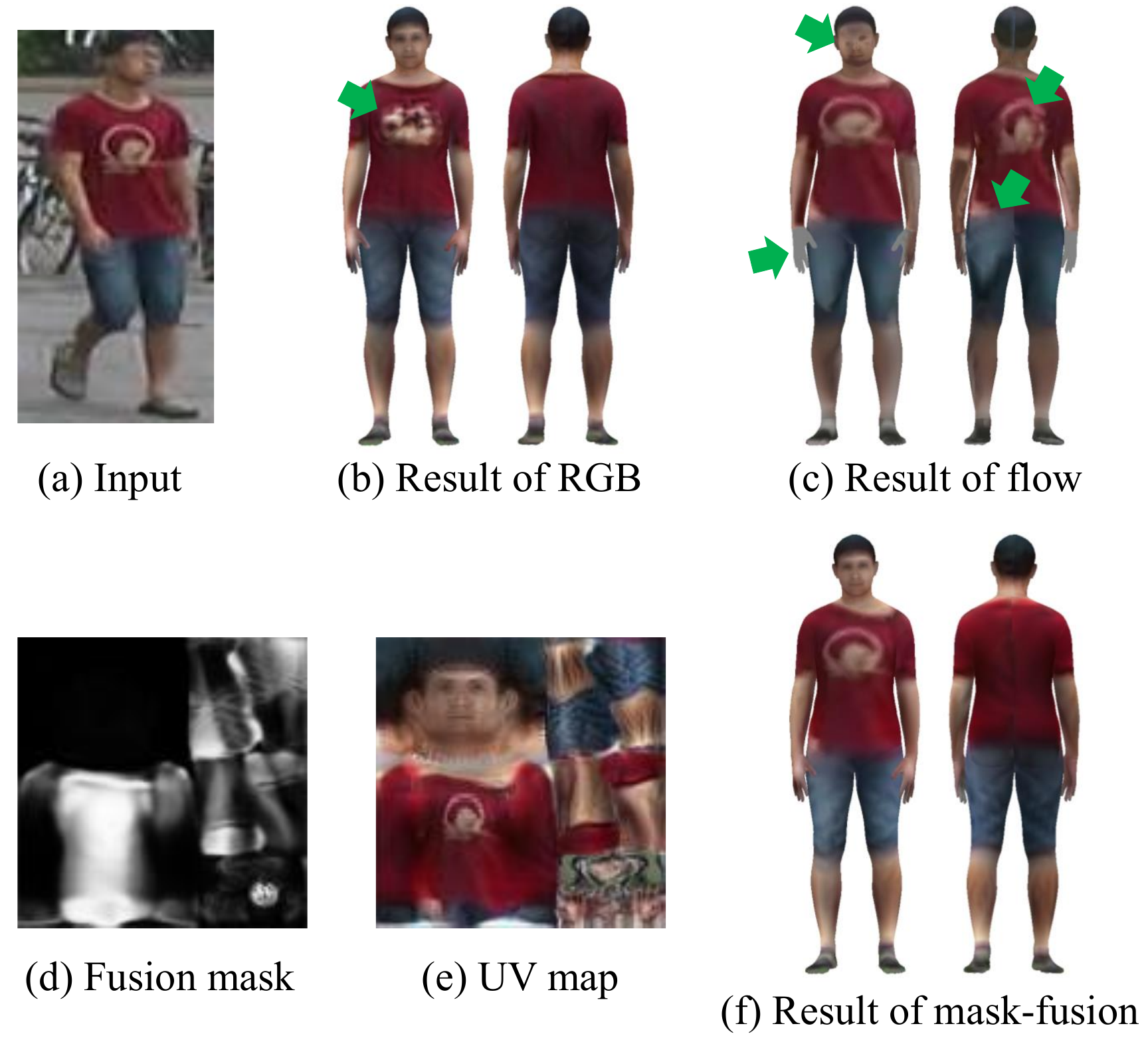} 
		\end{tabular}
	\end{center}
	\vspace{-3mm}
	\caption{Mask-fusion strategy combines RGB values and texture flow to reduce artifacts while preserving fine details.
	}
	\vspace{-5mm}
	\label{fig: rgb_flow}
\end{figure}

\noindent \textbf{Mask-fusion of RGB and texture flow.} 
There are two possible options for the output of the network: 1) directly generating the RGB textures $T_\text{RGB}$~\cite{wang2019re,xu2021_3d}, or 2) predicting the texture flow $F$ that can be used to generate the final textures by sampling from the input image $I$~\cite{kanazawa2018learning,zhao2020human,xu2019quadratic}.
As shown in Figure~\ref{fig: rgb_flow}, both these two strategies have their advantages and limitations:
Directly synthesizing $T_\text{RGB}$ can well reconstruct visually-pleasant 3D human textures, which however often leads to loss of fine details (Figure~\ref{fig: rgb_flow}(b)).
On the other hand, the texture flow is able to preserve fine details in the input image, but the results could suffer from severe artifacts (Figure~\ref{fig: rgb_flow}(c)).

To solve this problem, we propose a mask-fusion method to combine the advantages of both RGB values and texture flow.
Specifically, our Transformer generates three outputs: the RGB UV map $T_{\text{RGB}}$, the texture flow $F$, and a fusion mask $M$. 
Then the mask-fusion process can be written as:
\begin{align}
	T = M \odot f_\text{sample}(F, I) + (1-M) \odot T_{\text{RGB}},
\end{align}
where $f_\text{sample}$ is the bilinear sampling function that is used to sample textures from $I$ with $F$, and $\odot$ denotes element-wise multiplication.
The final texture $T$ is computed by fusing $T_{\text{RGB}}$ and $f_\text{sample}(F, I)$ via weighted sum.
As shown in Figure~\ref{fig: rgb_flow}(f), the mask-fusion method is able to significantly reduce visual artifacts while preserving fine details.
In addition, by observing the learned fusion mask (Figure~\ref{fig: rgb_flow}(d)), we can find that the network tends to use sampled textures $f_\text{sample}(F, I)$ for visible regions such as the human chest, and use synthesized textures $T_{\text{RGB}}$ for invisible regions such as the human back as well as other hard regions with sophisticated structures such as the face and hands.

\noindent \textbf{Relationship with conventional Transformers.}
Typically, existing works~\cite{vaswani2017attention,carion2020end,dosovitskiy2020image,chen2020pre} construct the Transformer by stacking multiple Transformer units, where the Output features of the current unit become the Query features for the next one.
Thus, the data flow in the Transformer needs to be able to serve as two different roles (\ie, Query and Output) simultaneously.
This is a reasonable choice when either 1) the Query and Output are in the same space, for example, in natural language processing~\cite{vaswani2017attention} the Query is the shifted Output of the previous time step, or 2) the physical meaning of the Query cannot be concretely defined, for instance, in object detection~\cite{carion2020end} the Query is randomly-initialized vectors that are also optimized during training.

However, in this work we have a clear definition of the Query which is a color encoding map obtained from 3D coordinates of a standard 3D body mesh, and each element in the Query corresponds to a physical vertex;
and this Query map is in a space different from the Output (\ie, the RGB values and texture flow).
Therefore, in our network (Figure~\ref{fig: overview}), we avoid stacking multiple attention modules and use the Transformer units more efficiently in a multi-scale manner.
Through this way, the features of our Texformer do not need to serve as two different roles simultaneously anymore, which effectively alleviates the difficulties of training.
As shown in Section~\ref{sec:exp_baselines}, the conventional Transformer cannot provide as high-quality results as the proposed model, which demonstrates the effectiveness of our design.

\subsection{Loss Functions} \label{sec: loss}
Following~\cite{wang2019re,xu2021_3d,kanazawa2018learning,zhao2020human}, we train the Texformer in a self-supervised manner.
First, we use the estimated human textures to render the human image $f_\text{r}(T, D)$, where $f_\text{r}$ is a differentiable rendering function~\cite{kato2018renderer}, and $D$ represents the 3D human mesh and camera parameters predicted by the state-of-the-art algorithm RSC-Net~\cite{xu20203d};
then the model can be trained by enforcing the similarity between the rendered image $f_\text{r}(T)$ and the input $I$.
Note that we omit $D$ here and in the following sections for conciseness.

To enforce the similarity between $f_\text{r}(T)$ and $I$, we first use the re-identification (ReID) loss from \cite{wang2019re}:
\begin{align}\label{eq: reid}
	\ell_{\text{ReID}} = \sum_{v} \| \bar{\phi}_v(f_\text{r}(T)) - \bar{\phi}_v(I) \|_2^2,
\end{align}
where $\phi_v$ represents the $v$-th layer of a pedestrian re-identification network~\cite{sun2018beyond}. 
$\bar{\phi}=\phi / \|\phi\|_2$ denotes L2-normalization.

\begin{figure}[t]
	\footnotesize
	\begin{center}
		\begin{tabular}{c}
			\includegraphics[width = 0.8\linewidth]{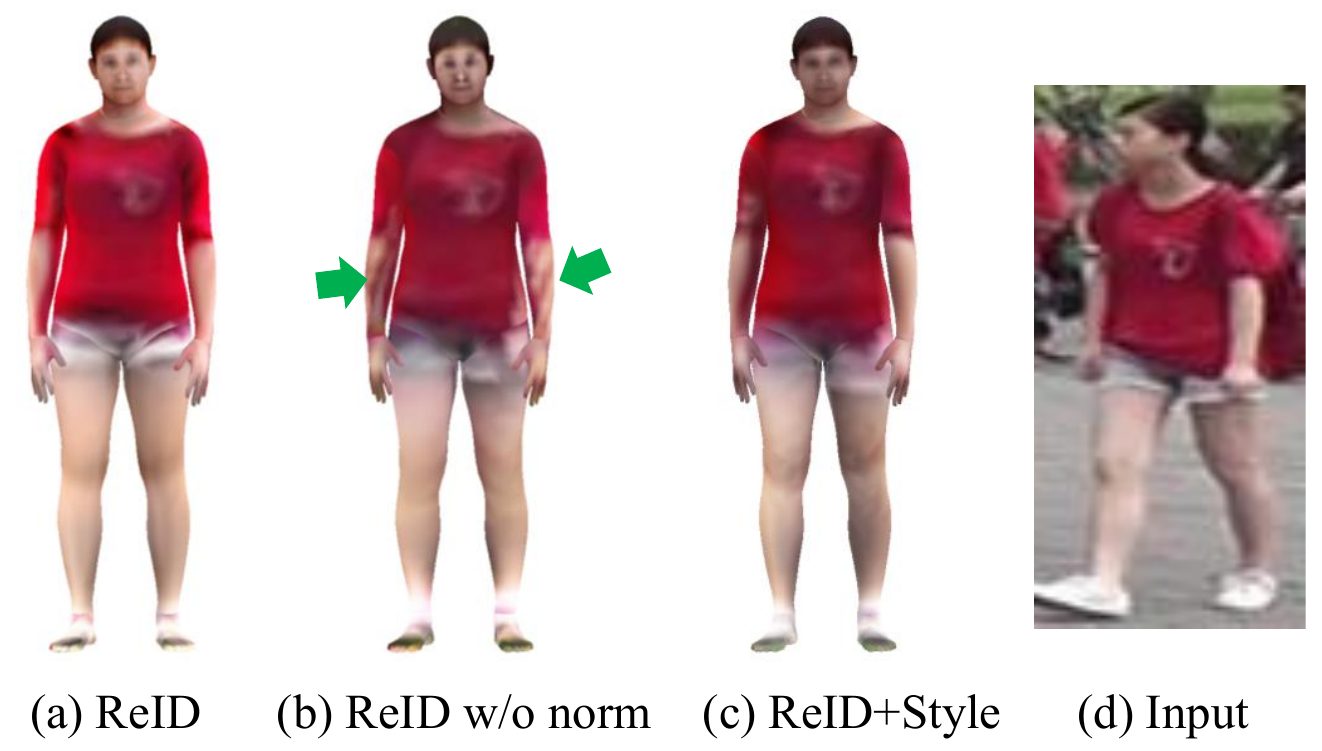} 
		\end{tabular}
	\end{center}
	\vspace{-3mm}
	\caption{Part-style loss for reconstructing human textures with high-fidelity colors.
	}
	\vspace{-3mm}
	\label{fig: style loss}
\end{figure}

While this ReID loss performs reasonably well, the reconstructed textures often have noticeable color differences from the input image as shown in Figure~\ref{fig: style loss}(a),
which is mainly due to the normalization operation in Eq.~\ref{eq: reid}.
One way to solve this problem is to remove the normalization and use the unnormalized features for the ReID loss.
However, this strategy often leads to severe artifacts in the reconstructed human textures as shown in Figure~\ref{fig: style loss}(b) (note the unnatural textures on human face and arms).

\begin{figure}[t]
	\footnotesize
	\begin{center}
		\begin{tabular}{c}
			\includegraphics[width = 0.8\linewidth]{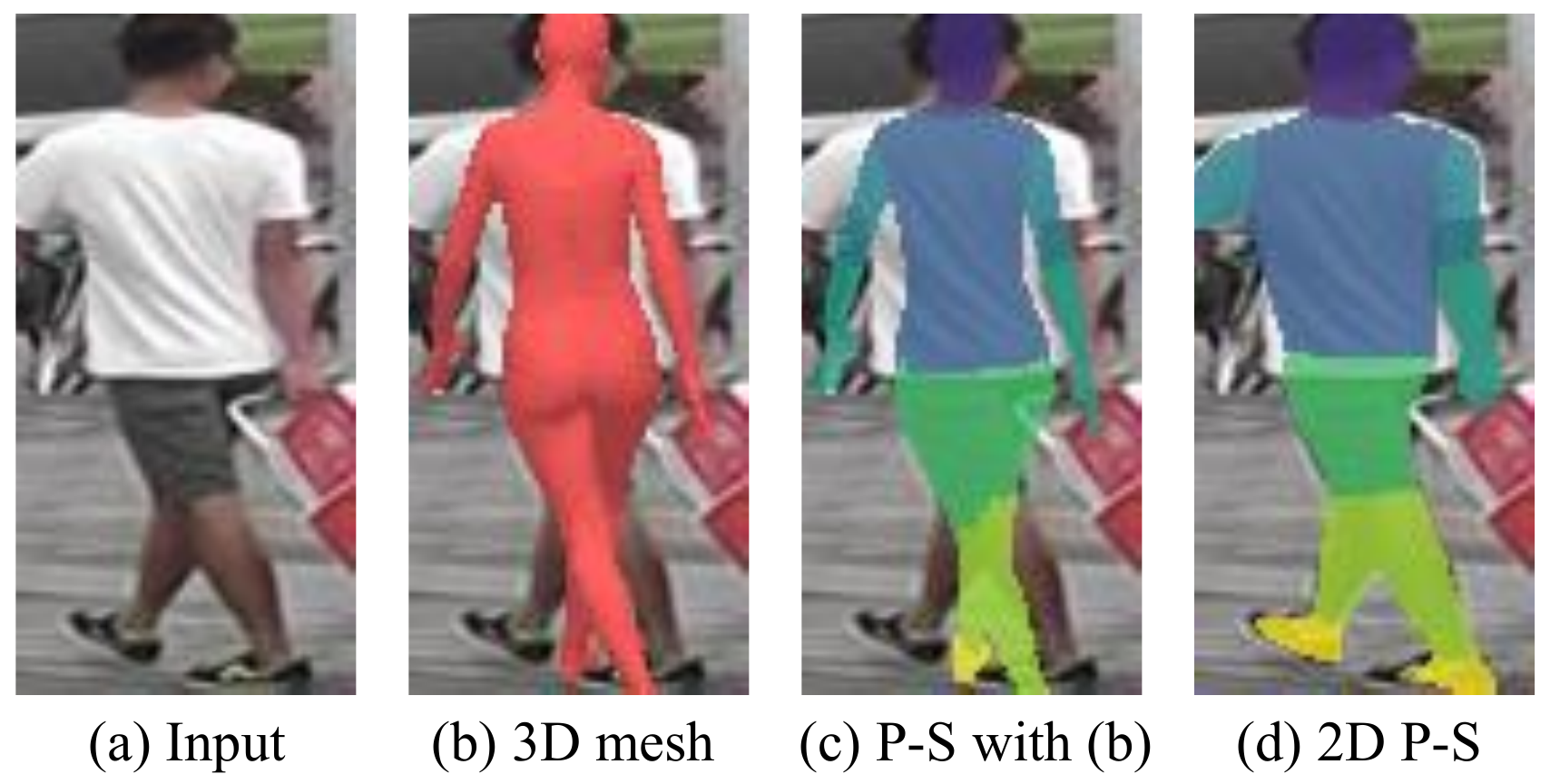} 
		\end{tabular}
	\end{center}
	\vspace{-2mm}
	\caption{Misalignment issue of the 3D mesh predictions.
		(b) is the 3D body mesh predicted by \cite{xu20203d}.
		(c) is the 2D body part segmentation (P-S) obtained from the 3D mesh (b).
		(d) is the 2D body part segmentation obtained by \cite{huang2018eanet}.
	}
	\vspace{-5mm}
	\label{fig: partseg}
\end{figure}

We suppose this problem could be owing to that the predicted 3D human mesh is not always accurate. 
As shown in Figure~\ref{fig: partseg}(b), the 3D human mesh does not align well with the input image. %
Therefore, directly enforcing the similarity between the rendered human and the input image leads to significant errors, which will negatively affect the training process.

\noindent \textbf{Part-style loss.} To address this issue, we propose a new loss function based on the observation that the 2D part-segmentation is generally more accurate than 3D estimation and aligns better with the input image (Figure~\ref{fig: partseg}(d)).
Specifically, we enforce the similarity between each body part of the rendered human and the input image, which naturally handles the misalignment issue caused by inaccurate 3D mesh predictions.

However, the regions of the same body part in different segmentation maps usually have different sizes and shapes as shown in Figure~\ref{fig: partseg}(c) and (d), which prevents us from using a simple MSE loss as in Eq.~\ref{eq: reid}. 
Instead, we employ the style loss from~\cite{gatys2016image} where the Gram matrix does not require the same size and shape for computation.
Different from the original style loss~\cite{gatys2016image} that computes the Gram matrix for the whole image, we enforce the Gram-matrix similarity in a per-body-part manner, which can be formulated as:
\begin{align}
	\ell_{\text{style}} = \sum_{p} \| G(M_p \odot \phi_1(f_\text{r}(T))) - G(M'_p \odot \phi_1(I)) \|_2^2, \nonumber
\end{align}
where $M$ and $M'$ are the human part segmentation from the 3D mesh and the 2D human parsing model~\cite{huang2018eanet}, respectively.
Here, $p$ indicates the $p$-th body part, and $G$ is the Gram matrix.
Note that we only use the features from the first layer of the ReID network (\ie, $\phi_1$) for the part-style loss to better focus on low-level colors of the reconstructed textures.
As shown in Figure~\ref{fig: style loss}(c), combining $\ell_\text{ReID}$ and $\ell_\text{style}$ achieves color appearances closer to the input image without introducing unpleasant artifacts.

\noindent \textbf{Face-structure loss.}
While our method is able to well reconstruct the 3D textures for most regions of the human body, it is still challenging to synthesize the textures for the human parts with complicated structures, such as the human face.
Wang~\etal~\cite{wang2019re} propose an MSE face loss to encourage the face textures to be close to the mean face of a synthetic human dataset~\cite{varol2017learning}, which however, often leads to unnatural results where the face color is not consistent with other skin regions of the reconstructed human.
To remedy this issue, we propose a face-structure loss:
\begin{align} \label{eq: face}
\ell_{\text{face}} = - \frac{1}{N} \sum_{i=1}^{N} s(M_\text{face} \odot T, M_\text{face} \odot T_\text{syn}^{(i)}),
\end{align}
where $\{T_\text{syn}^{(i)}\}_{i=1}^N$ is a collection of realistic human textures from a synthetic human dataset~\cite{varol2017learning}.
$M_\text{face}$ is a pre-defined binary map that indicates the face region on the texture map.
$s$ is a structure-similarity function~\cite{wang2004image} defined as:
\begin{align}
s(x,y)={(\sigma_{xy}+C)}/{(\sigma_x \sigma_y + C)},
\end{align}
where $\sigma_x$ is the standard deviation of $x$,
$\sigma_{xy}$ is the covariance between $x$ and $y$,
and $C$ is a constant.
Eq.~\ref{eq: face} essentially encourages the face textures to have similar structures as $T_\text{syn}$,
which effectively facilitates the generation of plausible face textures while retaining the colors of the input human.

Our final loss is a combination of the ReID loss, part-style loss, and face-structure loss:
\begin{align} \label{eq: final loss}
	\ell = w_1 * \ell_\text{ReID} + w_2 * \ell_\text{style} + w_3 * \ell_\text{face}, 
\end{align}
where $w_1$, $w_2$, and $w_3$ are hyper-parameters.
Similar to \cite{zhao2020human}, we use multi-view data to train the network, and our loss functions can be straightforwardly applied to multi-view images.

\begin{figure*}[t]
	\footnotesize
	\vspace{-9mm}
	\begin{center}
		\begin{tabular}{c}
			\includegraphics[width = 0.9\linewidth]{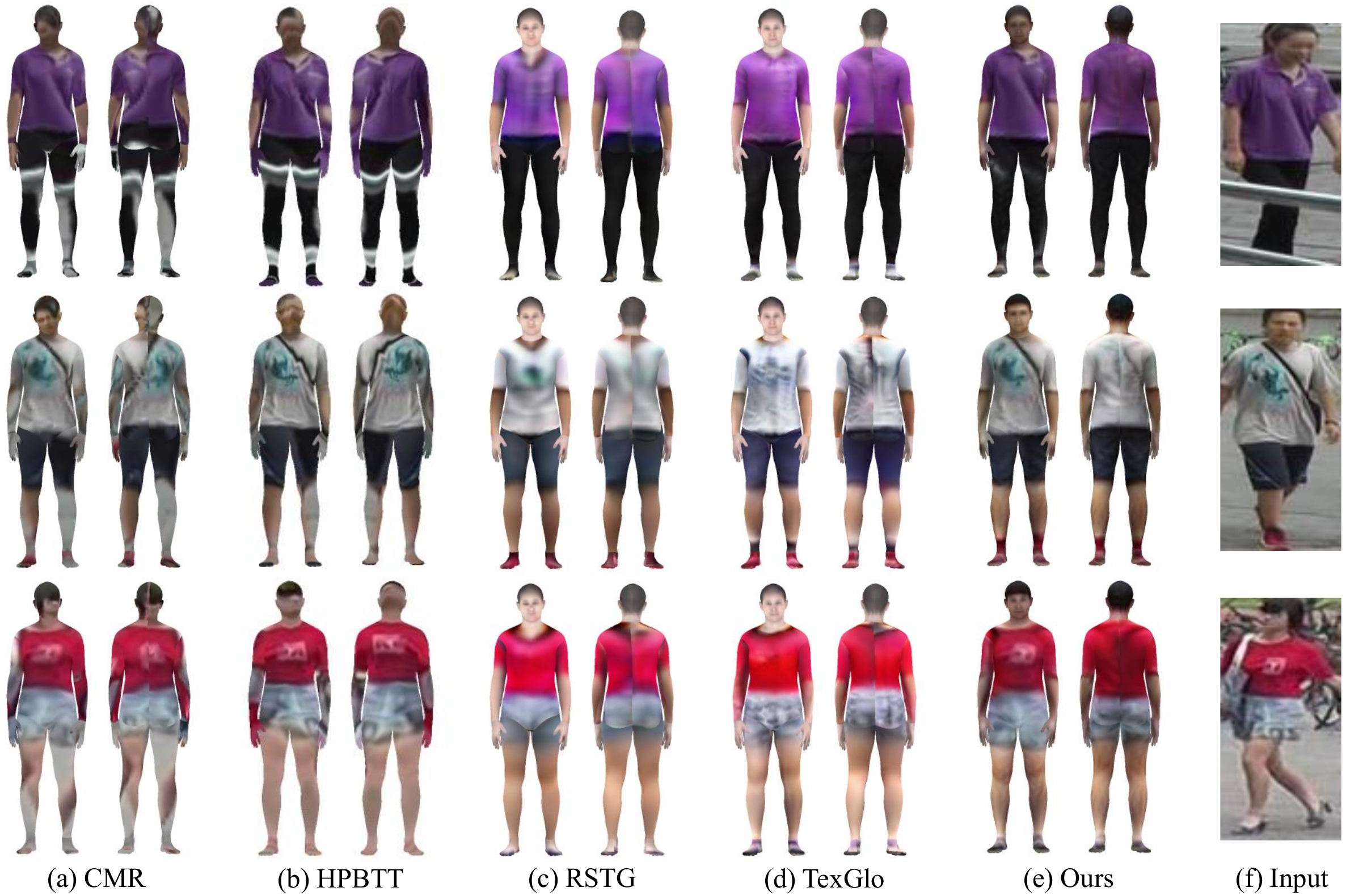} 
		\end{tabular}
	\end{center}
	\vspace{-3.5mm}
	\caption{Qualitative evaluation of the proposed algorithm.
	}
\vspace{-4.5mm}
	\label{fig: baselines}
\end{figure*}

\section{Experiments}
We use the Market-1501 dataset~\cite{zheng2015scalable} in our experiments, which consists of human images of 1501 person identities.
Following~\cite{wang2019re,xu2021_3d}, we use 1401 identities for training and the remaining 100 identities for testing.
For the face-structure loss, we use synthetic textures from the SURREAL dataset~\cite{varol2017learning}.
We use 8 attention heads for the attention modules, and the feature dimension is set as 128 for both $K$ and $V$. 
We set $\alpha=\tilde{h}\tilde{w}$ for the LoRA (Eq.~\ref{eq: low-rank-attn}).
We use BatchNorm~\cite{bnorm} instead of LayerNorm~\cite{ba2016layer} for the Transformer units as it gives better results in practice.
Similar to \cite{wang2019re,xu2021_3d}, we use the PCB network~\cite{sun2018beyond} as the feature extractor of the ReID loss (Eq.~\ref{eq: reid}).
We empirically set the coefficients in Eq.~\ref{eq: final loss} as $w_1=5000, w_2=0.4, w_3=0.01$.

For the evaluation metrics, we use the SSIM~\cite{wang2004image} and LPIPS~\cite{zhang2018unreasonable} with human masks~\cite{ma2017pose} to measure the quality of the predicted human textures. 
We also compute the cosine similarity (CosSim) of person ReID features~\cite{sun2018beyond} to measure the results from a semantic level. 
A higher CosSim indicates that the rendered human is more likely to be the same person in the input image.
As the PCB network~\cite{sun2018beyond} has been used for training the baselines and our model, we also compute the cosine similarity of the features from a ReID network unseen during training.
Specifically, we use the ResNet-50~\cite{resnet} from~\cite{zhou2019torchreid} and name the corresponding metric as CosSim-R.

\subsection{Comparison with the State of the Arts} \label{sec:exp_baselines}
We compare against the state-of-the-art 3D texture estimation methods: CMR~\cite{kanazawa2018learning}, HPBTT~\cite{zhao2020human}, RSTG~\cite{wang2019re}, and TexGlo~\cite{xu2021_3d}.
As HPBTT splits data in a different way, we retrain it with our protocol following \cite{wang2019re,xu2021_3d}.
As shown in Table~\ref{tab: sota}, the proposed method achieves consistently better results than the baseline approaches on all metrics while requiring a smaller amount of parameters, which demonstrates the effectiveness of our algorithm.

In addition, we also provide qualitative evaluations of our method against the baselines in Figure~\ref{fig: baselines}.
As the textures of CMR~\cite{kanazawa2018learning} and HPBTT~\cite{zhao2020human} are sampled from the input image with texture flow, they usually have colors close to the input, which leads to good performance in terms of SSIM and LPIPS in Table~\ref{tab: sota}.
However, these methods often suffer from significant artifacts and are not robust to occlusions as shown in Figure~\ref{fig: baselines}(a)-(b).
On the other hand, while RSTG~\cite{wang2019re} (Figure~\ref{fig: baselines}(c)) and TexGlo~\cite{xu2021_3d} (Figure~\ref{fig: baselines}(d)) can well reconstruct human textures without severe visual artifacts,
their results usually lack fine details and have significant color differences from the input. 
In contrast, the proposed algorithm achieves higher-quality results with fine details and high-fidelity colors in Figure~\ref{fig: baselines}(e).

\begin{table}[t] 
	\centering
	\scriptsize
	\caption{Quantitative evaluation of the proposed algorithm. Numbers in bold indicate the best in each column, and underscored numbers indicate the second.}
	\label{tab: sota}
	\begin{tabular}{lccccc}
		\toprule
		Method & CosSim $\uparrow$ & CosSim-R $\uparrow$ & SSIM $\uparrow$  & LPIPS $\downarrow$   & Params (M) \\
		\midrule
		CMR~\cite{kanazawa2018learning} & 0.5241 & 0.4978 & 0.7142 & 0.1275   & 84.1       \\
		HPBTT~\cite{zhao2020human} & 0.5246 & 0.5027 & \second{0.7420} & \second{0.1168}   & 235.0      \\
		RSTG~\cite{wang2019re} & 0.5282 & 0.4924  & 0.6735 & 0.1778   & 13.4       \\	
		TexGlo~\cite{xu2021_3d} & 0.5408 & 0.5048  & 0.6658 & 0.1776  & 16.1       \\
		Texformer & \first{0.5747} & \first{0.5422} & \first{0.7422} & \first{0.1154}   & 7.6       \\
		\midrule
		DETR~\cite{carion2020end} & 0.4344 & 0.4064 & 0.6482 & 0.2409   & 41.6   \\
		DETR*~\cite{carion2020end} & \second{0.5632} & \second{0.5274} & 0.7133 & 0.1379  & 17.9    \\
		\bottomrule
	\end{tabular}
	\vspace{-4mm}
\end{table}

\begin{figure}[t]
	\footnotesize
	\vspace{-1.mm}
	\begin{center}
		\begin{tabular}{c}
			\includegraphics[width = 0.74\linewidth]{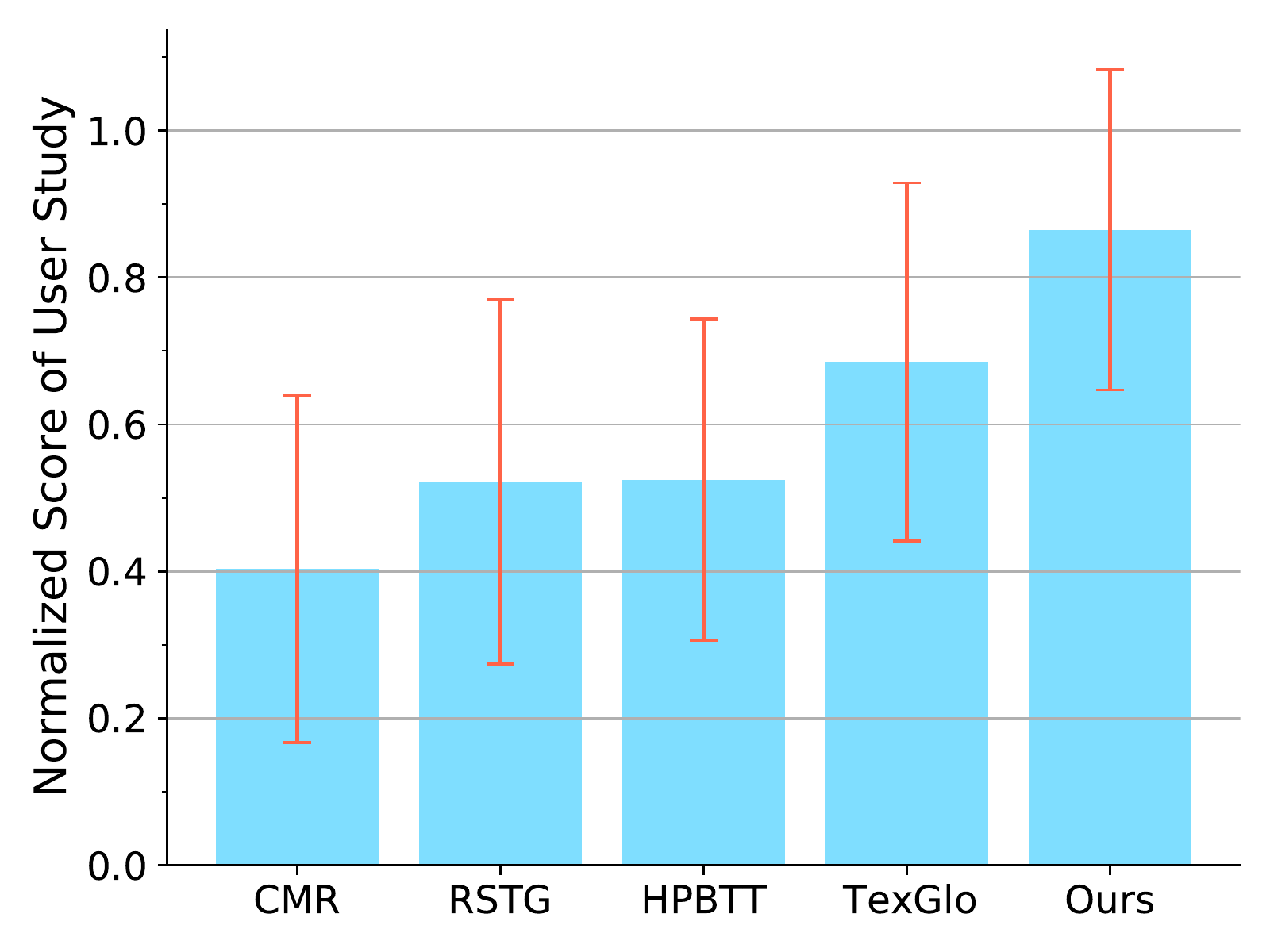} 
		\end{tabular}
	\end{center}
		\vspace{-3.5mm}
	\caption{User study of different algorithms for 3D human texture estimation. Each bar shows the mean and standard deviation of the normalized scores of each method.}
	\vspace{-5mm}
	\label{fig: user study}
\end{figure}

\noindent \textbf{User study.}
We conduct a user study for a more comprehensive evaluation of the texture estimation algorithms. This study uses 10 images randomly selected from the Market-1501 test set~\cite{zheng2015scalable}, and each input is processed by 5 different methods: CMR~\cite{kanazawa2018learning}, HPBTT~\cite{zhao2020human},  RSTG~\cite{wang2019re}, TexGlo~\cite{xu2021_3d}, and the proposed Texformer. 
Twenty subjects are asked to rank the reconstructed textures by each method with the input image as reference (1 for poor and 5 for excellent). 
We normalize the rank values to [0, 1], and use the normalized scores for measuring the results. 
We visualize the mean scores and the standard deviation in Figure~\ref{fig: user study}.
The proposed method is clearly preferred over other methods, suggesting its better capability in generating 3D human textures with high perceptual quality.

\begin{figure}[t]
	\footnotesize
	\begin{center}
		\begin{tabular}{c}
			\hspace{-1mm}
			\includegraphics[width = 0.75\linewidth]{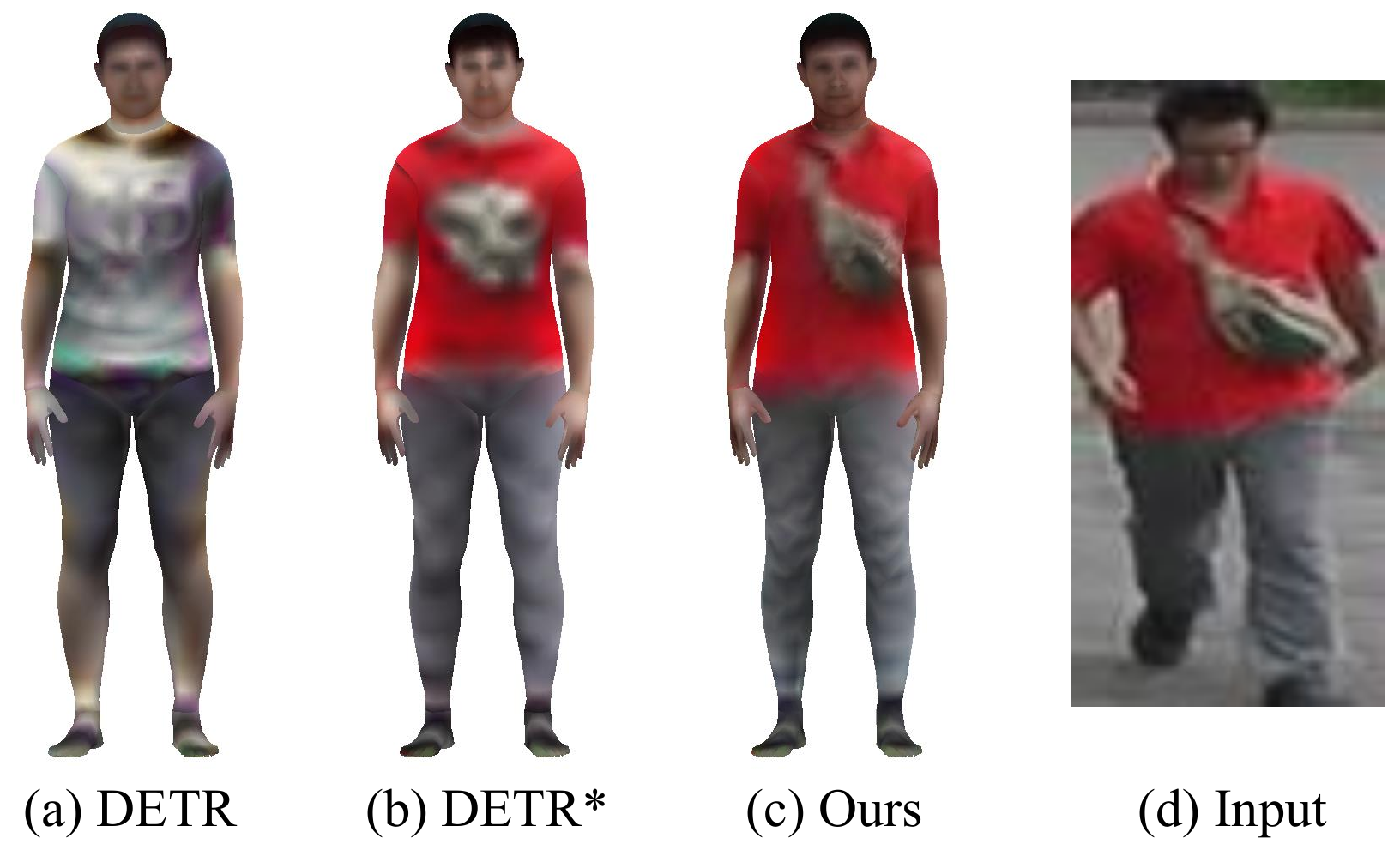} 
		\end{tabular}
	\end{center}
	\vspace{-3mm}
	\caption{Visual comparison with DETR~\cite{carion2020end}.}
	\vspace{-7mm}
	\label{fig: detr}
\end{figure}

\noindent \textbf{Comparison with DETR.}
The DETR~\cite{carion2020end}, which employs a conventional Transformer structure~\cite{vaswani2017attention}, is originally used for object detection. 
Here we adapt this network for 3D human texture estimation by replacing the object queries with UV map queries, which are randomly initialized and then optimized during training. 
We also add the proposed Query map to the queries of DETR for better performance. 
However, the original network cannot work properly as shown in Table~\ref{tab: sota} and Figure~\ref{fig: detr}(a), which is mainly due to the very deep structure (ResNet-50~\cite{resnet}) of DETR that prevents the preservation of lower-level texture information.
Therefore, we improve the original DETR by only using the first block of the ResNet-50 as the feature extractor.
This leads to a stronger baseline model (DETR*) as shown in Table~\ref{tab: sota} and Figure~\ref{fig: detr}(b).
However, as DETR stacks multiple attention modules together without explicitly disentangling the roles of the Query and Output features, it does not perform as well as the proposed network which produces high-quality human textures with fine details as shown in Figure~\ref{fig: detr}(c).  

\subsection{Ablation Study}
\begin{figure}[t]
	\footnotesize
	\vspace{-1mm}
	\begin{center}
		\begin{tabular}{c}
			\hspace{-3mm}
			\includegraphics[width = 0.99\linewidth]{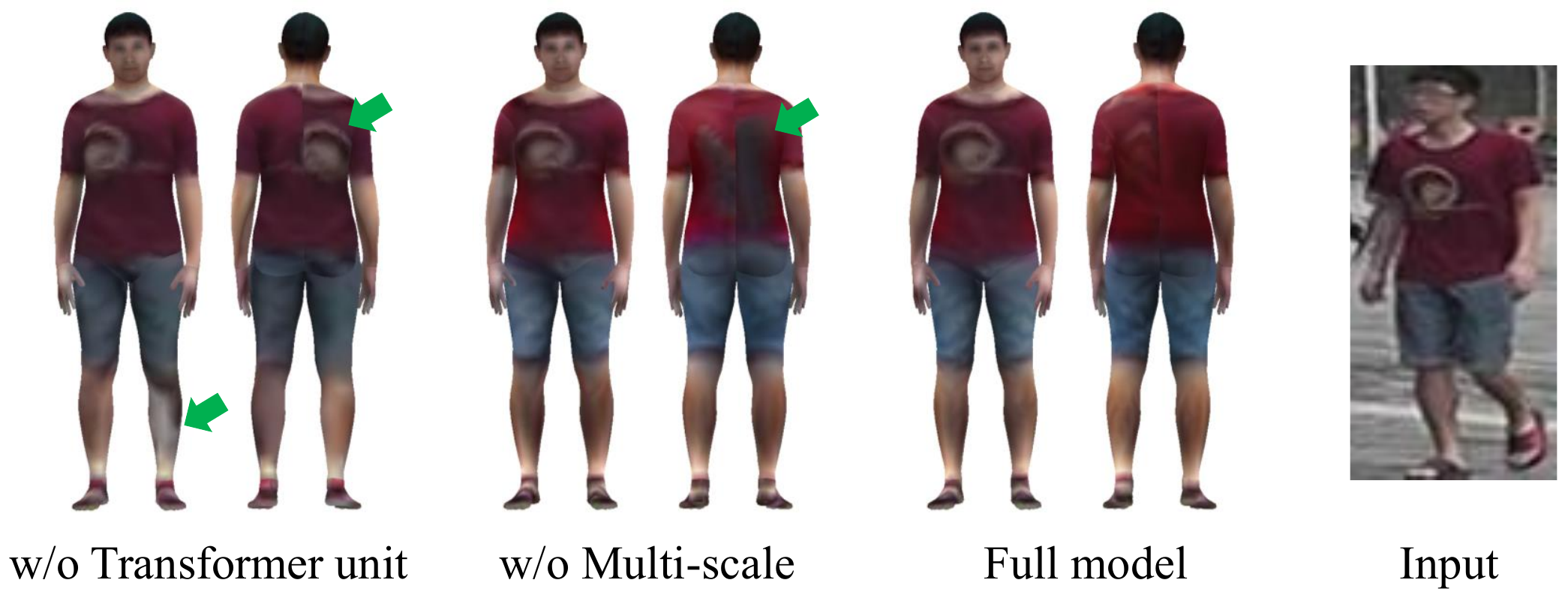} 
		\end{tabular}
	\end{center}
		\vspace{-3mm}
	\caption{Effectiveness of different components of Texformer.}
	\vspace{-4mm}
	\label{fig: ablation}
\end{figure}

\begin{figure}[t]
	\footnotesize
	\begin{center}
		\begin{tabular}{c}
			\includegraphics[width = 0.99\linewidth]{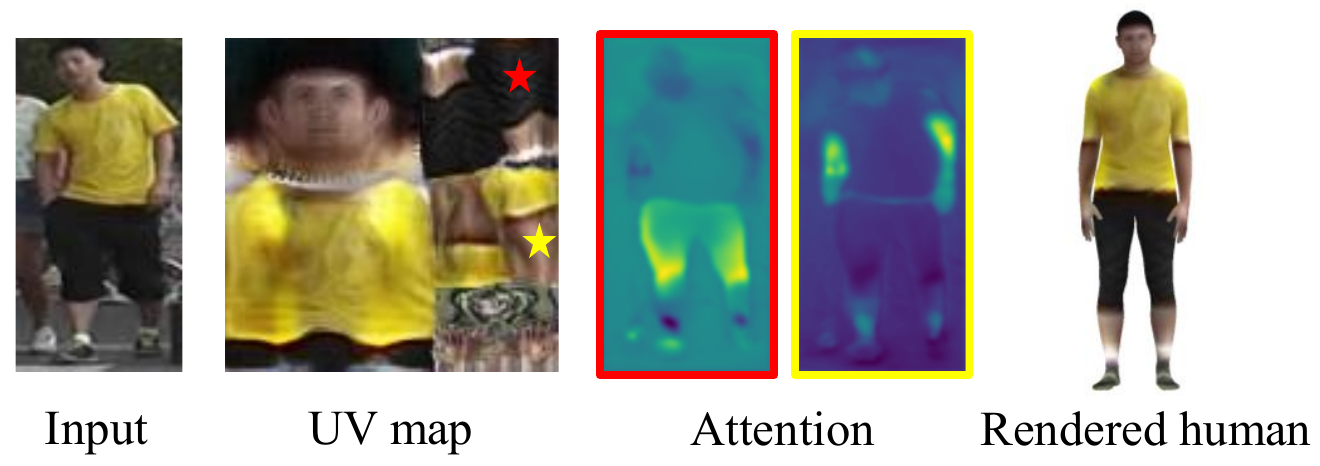} 
		\end{tabular}
	\end{center}
	\vspace{-3mm}
	\caption{Visualization of the learned attention map. The red and yellow stars in the UV map correspond to the attention maps surrounded by red and yellow frames, respectively.}
	\vspace{-5mm}
	\label{fig: attention}
\end{figure}

\noindent \textbf{Effectiveness of the Transformer.}
The proposed network essentially relies on the Transformer unit (Eq.~\ref{eq: trans unit}) to exploit the global information of the input.
To analyze its effect, we remove the Transformer units in Figure~\ref{fig: overview} and instead generate the Output features by first concatenating $Q$, $K$, and $V$ and then sending the concatenated features into a normal convolution layer.
As shown in Table~\ref{tab: ablation}, the model without the Transformer unit suffers from a significant performance drop in terms of all the metrics. %
We also provide a visual example in Figure~\ref{fig: ablation}, where the result of ``w/o Transformer unit'' has much lower visual quality with unpleasant artifacts.
To better understand the attention mechanism of the Transformer unit, we further present a visualization of the learned attention map in Figure~\ref{fig: attention},
showing that the Transformer unit can learn to aggregate relevant features from the image into the UV space.

Furthermore, in our approach we use a multi-scale strategy to fuse the features from the Transformer units, which moderately improves the performance as shown in Table~\ref{tab: ablation}.
As the multi-scale strategy can effectively exploit the contextual information of the input, it is able to better infer textures for invisible regions of the human body, \eg the human back in Figure~\ref{fig: ablation}.

\begin{table}[]
	\centering
\scriptsize
	\caption{Ablation study of the proposed method.}
	\label{tab: ablation}
	\begin{tabular}{lcccc}
		\toprule
		Method  & CosSim $\uparrow$ & CosSim-R $\uparrow$ & SSIM $\uparrow$   & LPIPS $\downarrow$ \\
		\midrule
		w/o Transformer unit & 0.5413 & 0.5101   & 0.7242 & 0.1235       \\
		w/o multi-scale & 0.5715 & 0.5391     & {0.7407} & {0.1181}      \\
		\midrule
		only RGB & {0.5711} & {0.5370} & 0.7167 & 0.1261 \\
		only texture flow & 0.5515 & 0.5258 & \first{0.7581} & \first{0.1027} \\
		\midrule
		w/o part-style loss & \first{0.5783} & \first{0.5441} & 0.7158 & 0.1412  \\
		\midrule
		Full model & \second{0.5747} & \second{0.5422}    & \second{0.7422} & \second{0.1154}     \\
		\bottomrule
	\end{tabular}
\vspace{-1.5mm}
\end{table}

\noindent \textbf{Effectiveness of combining RGB and texture flow.}
As introduced in Section~\ref{sec: transformer}, we propose a mask-fusion method to combine the RGB output and texture flow.
As shown in Table~\ref{tab: ablation}, directly predicting the RGB values leads to poor SSIM and LPIPS, which is mainly due to the incapability of the RGB-based model to reconstruct accurate details (Figure~\ref{fig: rgb_flow}(b)).
Furthermore, only using texture flow for human texture estimation results in low CosSim and CosSim-R in Table~\ref{tab: ablation}, which is largely caused by the significant amount of artifacts as in Figure~\ref{fig: rgb_flow}(c). 
Note that while both our RGB-based and flow-based models have their limitations, they still achieve better results than the previous RGB-based and flow-based approaches (Table~\ref{tab: sota}), which again shows the superiority of the proposed Transformer network.

\begin{figure}[t]
	\footnotesize
	\vspace{-1mm}
	\begin{center}
		\begin{tabular}{c}
			\includegraphics[width = 0.89\linewidth]{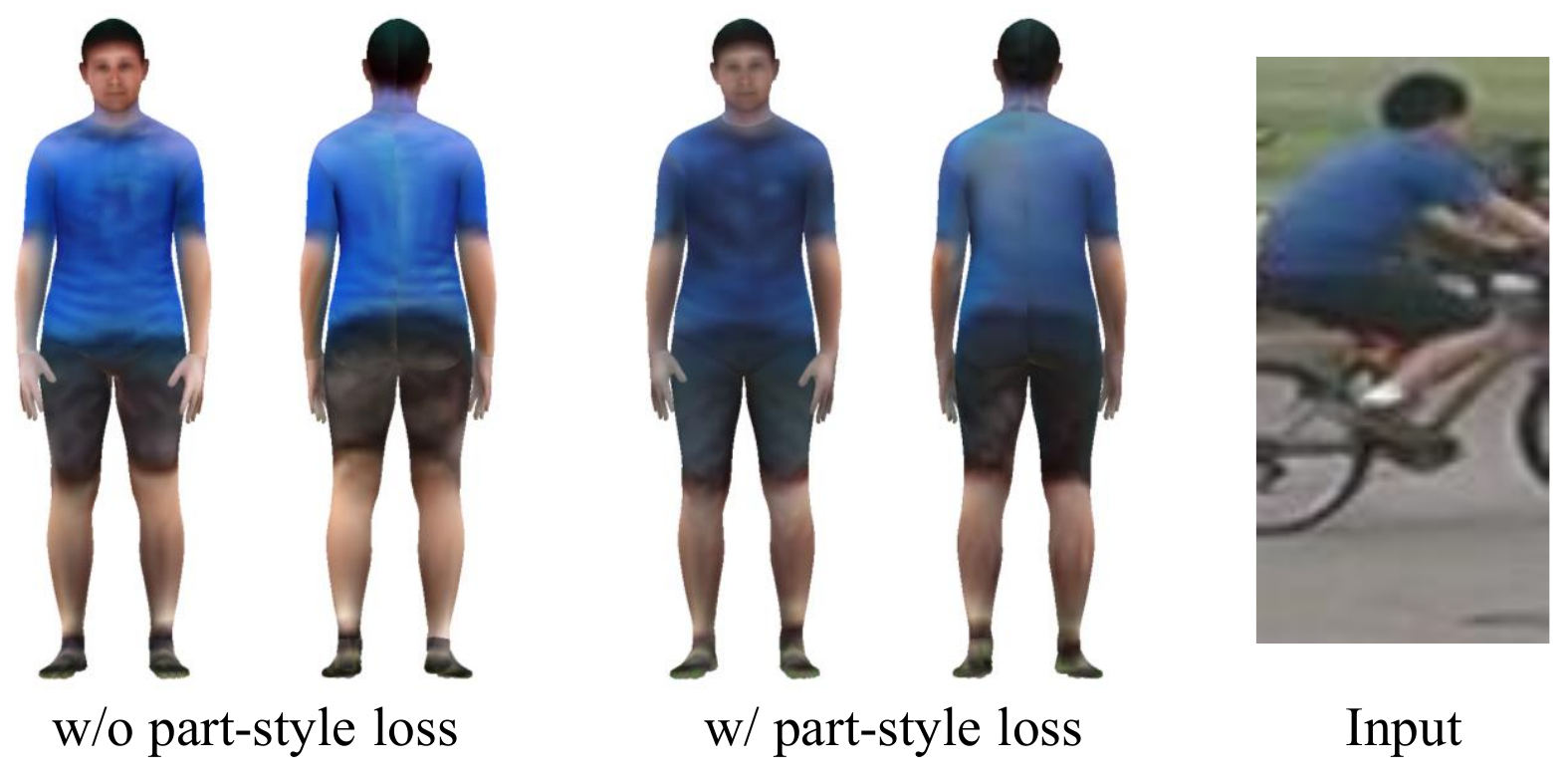} 
		\end{tabular}
	\end{center}
	\vspace{-3.5mm}
	\caption{Effectiveness of the part-style loss.
	}
	\vspace{-5.5mm}
	\label{fig: style loss2}
\end{figure}

\noindent \textbf{Effectiveness of the part-style loss.}
As shown in Table~\ref{tab: ablation}, the part-style loss significantly improves the SSIM and LPIPS. 
In Figure~\ref{fig: style loss2}, the reconstructed textures without this loss have noticeable color differences from the input image.
This is consistent with the motivation of our design in Section~\ref{sec: loss} to synthesize high-fidelity colors without introducing unpleasant visual artifacts.
Note that the part-style loss leads to a slight decrease of the CosSim and CosSim-R in Table~\ref{tab: ablation}. 
This is possibly due to that the ReID network does not fully rely on color appearances to recognize humans, as the same person may appear in different colors when viewed from different angles or under different light conditions.
Therefore, enforcing the output to have close color appearances to the input may not always coincide with the optimization direction of the ReID loss, and may slightly distract the training process.

\section{Concluding Remarks}
We have developed the Texformer for high-quality 3D human texture reconstruction from a single image. 
At the core of this work is the Transformer unit that allows efficient information interchange between two different spaces, \ie the image space and UV texture space.
Potentially, this idea could be  extended to other fields where the data of interest involves multiple representations or lies in different spaces, such as multi-modality learning and graph processing. 

On the other hand, due to the limitation of SMPL~\cite{loper2015smpl}, the proposed method is mostly suitable for tight-fitting clothes rather than loose-fitting ones. 
It is possible to overcome this limitation by introducing more advanced human body models in Texformer, which could be an interesting direction for future research.

\noindent \textbf{Acknowledgement.} This study is supported under the RIE2020 Industry Alignment Fund - Industry Collaboration Projects (IAF-ICP) Funding Initiative, as well as cash and in-kind contribution from the industry partner(s).

{\small
\bibliographystyle{ieee_fullname}
\bibliography{egbib}

\begin{thebibliography}{10}\itemsep=-1pt

\bibitem{alldieck2019learning}
Thiemo Alldieck, Marcus Magnor, Bharat~Lal Bhatnagar, Christian Theobalt, and
  Gerard Pons-Moll.
\newblock Learning to reconstruct people in clothing from a single rgb camera.
\newblock In {\em CVPR}, 2019.

\bibitem{alldieck2018detailed}
Thiemo Alldieck, Marcus Magnor, Weipeng Xu, Christian Theobalt, and Gerard
  Pons-Moll.
\newblock Detailed human avatars from monocular video.
\newblock In {\em International Conference on 3D Vision (3DV)}, 2018.

\bibitem{alldieck2018video}
Thiemo Alldieck, Marcus Magnor, Weipeng Xu, Christian Theobalt, and Gerard
  Pons-Moll.
\newblock Video based reconstruction of 3d people models.
\newblock In {\em CVPR}, 2018.

\bibitem{alldieck2019tex2shape}
Thiemo Alldieck, Gerard Pons-Moll, Christian Theobalt, and Marcus Magnor.
\newblock Tex2shape: Detailed full human body geometry from a single image.
\newblock In {\em ICCV}, 2019.

\bibitem{ba2016layer}
Jimmy~Lei Ba, Jamie~Ryan Kiros, and Geoffrey~E Hinton.
\newblock Layer normalization.
\newblock {\em arXiv preprint arXiv:1607.06450}, 2016.

\bibitem{bhatnagar2019multi}
Bharat~Lal Bhatnagar, Garvita Tiwari, Christian Theobalt, and Gerard Pons-Moll.
\newblock Multi-garment net: Learning to dress {3D} people from images.
\newblock In {\em ICCV}, 2019.

\bibitem{boykov2001fast}
Yuri Boykov, Olga Veksler, and Ramin Zabih.
\newblock Fast approximate energy minimization via graph cuts.
\newblock {\em TPAMI}, 23(11):1222--1239, 2001.

\bibitem{carion2020end}
Nicolas Carion, Francisco Massa, Gabriel Synnaeve, Nicolas Usunier, Alexander
  Kirillov, and Sergey Zagoruyko.
\newblock End-to-end object detection with transformers.
\newblock In {\em ECCV}, 2020.

\bibitem{chen2020pre}
Hanting Chen, Yunhe Wang, Tianyu Guo, Chang Xu, Yiping Deng, Zhenhua Liu, Siwei
  Ma, Chunjing Xu, Chao Xu, and Wen Gao.
\newblock Pre-trained image processing transformer.
\newblock {\em arXiv preprint arXiv:2012.00364}, 2020.

\bibitem{devlin2018bert}
Jacob Devlin, Ming-Wei Chang, Kenton Lee, and Kristina Toutanova.
\newblock Bert: Pre-training of deep bidirectional transformers for language
  understanding.
\newblock {\em arXiv preprint arXiv:1810.04805}, 2018.

\bibitem{dosovitskiy2020image}
Alexey Dosovitskiy, Lucas Beyer, Alexander Kolesnikov, Dirk Weissenborn,
  Xiaohua Zhai, Thomas Unterthiner, Mostafa Dehghani, Matthias Minderer, Georg
  Heigold, Sylvain Gelly, et~al.
\newblock An image is worth 16x16 words: Transformers for image recognition at
  scale.
\newblock {\em arXiv preprint arXiv:2010.11929}, 2020.

\bibitem{gatys2016image}
Leon~A Gatys, Alexander~S Ecker, and Matthias Bethge.
\newblock Image style transfer using convolutional neural networks.
\newblock In {\em CVPR}, 2016.

\bibitem{goel2020shape}
Shubham Goel, Angjoo Kanazawa, and Jitendra Malik.
\newblock Shape and viewpoint without keypoints.
\newblock In {\em ECCV}, 2020.

\bibitem{han2020survey}
Kai Han, Yunhe Wang, Hanting Chen, Xinghao Chen, Jianyuan Guo, Zhenhua Liu,
  Yehui Tang, An Xiao, Chunjing Xu, Yixing Xu, et~al.
\newblock A survey on visual transformer.
\newblock {\em arXiv preprint arXiv:2012.12556}, 2020.

\bibitem{resnet}
Kaiming He, Xiangyu Zhang, Shaoqing Ren, and Jian Sun.
\newblock Deep residual learning for image recognition.
\newblock In {\em CVPR}, 2016.

\bibitem{huang2018eanet}
Houjing Huang, Wenjie Yang, Xiaotang Chen, Xin Zhao, Kaiqi Huang, Jinbin Lin,
  Guan Huang, and Dalong Du.
\newblock Eanet: Enhancing alignment for cross-domain person re-identification.
\newblock {\em arXiv preprint arXiv:1812.11369}, 2018.

\bibitem{huang2020hand}
Lin Huang, Jianchao Tan, Ji Liu, and Junsong Yuan.
\newblock Hand-transformer: Non-autoregressive structured modeling for {3D}
  hand pose estimation.
\newblock In {\em ECCV}, 2020.

\bibitem{huang2018deep}
Zeng Huang, Tianye Li, Weikai Chen, Yajie Zhao, Jun Xing, Chloe LeGendre,
  Linjie Luo, Chongyang Ma, and Hao Li.
\newblock Deep volumetric video from very sparse multi-view performance
  capture.
\newblock In {\em ECCV}, 2018.

\bibitem{huang2020arch}
Zeng Huang, Yuanlu Xu, Christoph Lassner, Hao Li, and Tony Tung.
\newblock Arch: Animatable reconstruction of clothed humans.
\newblock In {\em CVPR}, 2020.

\bibitem{bnorm}
Sergey Ioffe and Christian Szegedy.
\newblock Batch normalization: Accelerating deep network training by reducing
  internal covariate shift.
\newblock In {\em ICML}, 2015.

\bibitem{isola2017image}
Phillip Isola, Jun-Yan Zhu, Tinghui Zhou, and Alexei~A Efros.
\newblock Image-to-image translation with conditional adversarial networks.
\newblock In {\em CVPR}, 2017.

\bibitem{kanazawa2018end}
Angjoo Kanazawa, Michael~J Black, David~W Jacobs, and Jitendra Malik.
\newblock End-to-end recovery of human shape and pose.
\newblock In {\em CVPR}, 2018.

\bibitem{kanazawa2018learning}
Angjoo Kanazawa, Shubham Tulsiani, Alexei~A Efros, and Jitendra Malik.
\newblock Learning category-specific mesh reconstruction from image
  collections.
\newblock In {\em ECCV}, 2018.

\bibitem{kanazawa2019learning}
Angjoo Kanazawa, Jason~Y Zhang, Panna Felsen, and Jitendra Malik.
\newblock Learning 3d human dynamics from video.
\newblock In {\em CVPR}, 2019.

\bibitem{kato2018renderer}
Hiroharu Kato, Yoshitaka Ushiku, and Tatsuya Harada.
\newblock Neural {3D} mesh renderer.
\newblock In {\em CVPR}, 2018.

\bibitem{kocabas2019vibe}
Muhammed Kocabas, Nikos Athanasiou, and Michael~J Black.
\newblock Vibe: Video inference for human body pose and shape estimation.
\newblock In {\em CVPR}, 2020.

\bibitem{kolotouros2019spin}
Nikos Kolotouros, Georgios Pavlakos, Michael~J. Black, and Kostas Daniilidis.
\newblock Learning to reconstruct {3D} human pose and shape via model-fitting
  in the loop.
\newblock In {\em ICCV}, 2019.

\bibitem{lazova2019360}
Verica Lazova, Eldar Insafutdinov, and Gerard Pons-Moll.
\newblock 360-degree textures of people in clothing from a single image.
\newblock In {\em International Conference on 3D Vision (3DV)}, 2019.

\bibitem{li2020self}
Xueting Li, Sifei Liu, Kihwan Kim, Shalini De~Mello, Varun Jampani, Ming-Hsuan
  Yang, and Jan Kautz.
\newblock Self-supervised single-view {3D} reconstruction via semantic
  consistency.
\newblock In {\em ECCV}, 2020.

\bibitem{li2020robust}
Zhe Li, Tao Yu, Chuanyu Pan, Zerong Zheng, and Yebin Liu.
\newblock Robust 3d self-portraits in seconds.
\newblock In {\em CVPR}, 2020.

\bibitem{loper2015smpl}
Matthew Loper, Naureen Mahmood, Javier Romero, Gerard Pons-Moll, and Michael~J
  Black.
\newblock Smpl: A skinned multi-person linear model.
\newblock {\em ACM Transactions on Graphics}, 34(6):248, 2015.

\bibitem{ma2017pose}
Liqian Ma, Xu Jia, Qianru Sun, Bernt Schiele, Tinne Tuytelaars, and Luc
  Van~Gool.
\newblock Pose guided person image generation.
\newblock In {\em NeurIPS}, 2017.

\bibitem{maneewongvatana1999analysis}
Songrit Maneewongvatana and David~M Mount.
\newblock Analysis of approximate nearest neighbor searching with clustered
  point sets.
\newblock In {\em ALENEX}, 1999.

\bibitem{mir2020learning}
Aymen Mir, Thiemo Alldieck, and Gerard Pons-Moll.
\newblock Learning to transfer texture from clothing images to {3D} humans.
\newblock In {\em CVPR}, 2020.

\bibitem{natsume2019siclope}
Ryota Natsume, Shunsuke Saito, Zeng Huang, Weikai Chen, Chongyang Ma, Hao Li,
  and Shigeo Morishima.
\newblock Siclope: Silhouette-based clothed people.
\newblock In {\em CVPR}, 2019.

\bibitem{neverova2018dense}
Natalia Neverova, Riza~Alp Guler, and Iasonas Kokkinos.
\newblock Dense pose transfer.
\newblock In {\em ECCV}, 2018.

\bibitem{pumarola20193dpeople}
Albert Pumarola, Jordi Sanchez-Riera, Gary Choi, Alberto Sanfeliu, and Francesc
  Moreno-Noguer.
\newblock {3DPeople}: Modeling the geometry of dressed humans.
\newblock In {\em ICCV}, 2019.

\bibitem{ronneberger2015u}
Olaf Ronneberger, Philipp Fischer, and Thomas Brox.
\newblock U-net: Convolutional networks for biomedical image segmentation.
\newblock In {\em International Conference on Medical image computing and
  computer-assisted intervention}, 2015.

\bibitem{saito2019pifu}
Shunsuke Saito, Zeng Huang, Ryota Natsume, Shigeo Morishima, Angjoo Kanazawa,
  and Hao Li.
\newblock Pifu: Pixel-aligned implicit function for high-resolution clothed
  human digitization.
\newblock In {\em ICCV}, 2019.

\bibitem{saito2020pifuhd}
Shunsuke Saito, Tomas Simon, Jason Saragih, and Hanbyul Joo.
\newblock Pifuhd: Multi-level pixel-aligned implicit function for
  high-resolution {3D} human digitization.
\newblock In {\em CVPR}, 2020.

\bibitem{sun2018beyond}
Yifan Sun, Liang Zheng, Yi Yang, Qi Tian, and Shengjin Wang.
\newblock Beyond part models: Person retrieval with refined part pooling (and a
  strong convolutional baseline).
\newblock In {\em ECCV}, 2018.

\bibitem{varol2017learning}
Gul Varol, Javier Romero, Xavier Martin, Naureen Mahmood, Michael~J Black, Ivan
  Laptev, and Cordelia Schmid.
\newblock Learning from synthetic humans.
\newblock In {\em CVPR}, 2017.

\bibitem{vaswani2017attention}
Ashish Vaswani, Noam Shazeer, Niki Parmar, Jakob Uszkoreit, Llion Jones,
  Aidan~N Gomez, Lukasz Kaiser, and Illia Polosukhin.
\newblock Attention is all you need.
\newblock In {\em NIPS}, 2017.

\bibitem{wang2019re}
Jian Wang, Yunshan Zhong, Yachun Li, Chi Zhang, and Yichen Wei.
\newblock Re-identification supervised texture generation.
\newblock In {\em CVPR}, 2019.

\bibitem{wang2004image}
Zhou Wang, Alan~C Bovik, Hamid~R Sheikh, and Eero~P Simoncelli.
\newblock Image quality assessment: from error visibility to structural
  similarity.
\newblock {\em TIP}, 13(4):600--612, 2004.

\bibitem{xu20203d}
Xiangyu Xu, Hao Chen, Francesc Moreno-Noguer, Laszlo~A Jeni, and Fernando De~la
  Torre.
\newblock {3D} human shape and pose from a single low-resolution image with
  self-supervised learning.
\newblock In {\em ECCV}, 2020.

\bibitem{xu2021_3d}
Xiangyu Xu, Hao Chen, Francesc Moreno-Noguer, Laszlo~A Jeni, and Fernando De~la
  Torre.
\newblock {3D} human pose, shape and texture from low-resolution images and
  videos.
\newblock {\em TPAMI}, 2021.

\bibitem{xu2020learning_stpan}
Xiangyu Xu, Muchen Li, Wenxiu Sun, and Ming-Hsuan Yang.
\newblock Learning spatial and spatio-temporal pixel aggregations for image and
  video denoising.
\newblock {\em TIP}, 29:7153--7165, 2020.

\bibitem{xu2019quadratic}
Xiangyu Xu, Siyao Li, Wenxiu Sun, Qian Yin, and Ming-Hsuan Yang.
\newblock Quadratic video interpolation.
\newblock In {\em NeurIPS}, 2019.

\bibitem{xu2020learning}
Xiangyu Xu, Yongrui Ma, and Wenxiu Sun.
\newblock Learning factorized weight matrix for joint filtering.
\newblock In {\em ICML}, 2020.

\bibitem{xxy-iccv17}
Xiangyu Xu, Deqing Sun, Jinshan Pan, Yujin Zhang, Hanspeter Pfister, and
  Ming-Hsuan Yang.
\newblock Learning to super-resolve blurry face and text images.
\newblock In {\em ICCV}, 2017.

\bibitem{zhang2018unreasonable}
Richard Zhang, Phillip Isola, Alexei~A Efros, Eli Shechtman, and Oliver Wang.
\newblock The unreasonable effectiveness of deep features as a perceptual
  metric.
\newblock In {\em CVPR}, 2018.

\bibitem{zhao2020human}
Fang Zhao, Shengcai Liao, Kaihao Zhang, and Ling Shao.
\newblock Human parsing based texture transfer from single image to {3D} human
  via cross-view consistency.
\newblock In {\em NeurIPS}, 2020.

\bibitem{zheng2015scalable}
Liang Zheng, Liyue Shen, Lu Tian, Shengjin Wang, Jingdong Wang, and Qi Tian.
\newblock Scalable person re-identification: A benchmark.
\newblock In {\em ICCV}, 2015.

\bibitem{zheng2020pamir}
Zerong Zheng, Tao Yu, Yebin Liu, and Qionghai Dai.
\newblock {PaMIR}: Parametric model-conditioned implicit representation for
  image-based human reconstruction.
\newblock {\em TPAMI}, 2021.

\bibitem{zheng2019deephuman}
Zerong Zheng, Tao Yu, Yixuan Wei, Qionghai Dai, and Yebin Liu.
\newblock Deephuman: {3D} human reconstruction from a single image.
\newblock In {\em ICCV}, 2019.

\bibitem{zhi2020texmesh}
Tiancheng Zhi, Christoph Lassner, Tony Tung, Carsten Stoll, Srinivasa~G
  Narasimhan, and Minh Vo.
\newblock Texmesh: Reconstructing detailed human texture and geometry from
  rgb-d video.
\newblock In {\em ECCV}, 2020.

\bibitem{zhou2019torchreid}
Kaiyang Zhou and Tao Xiang.
\newblock Torchreid: A library for deep learning person re-identification in
  pytorch.
\newblock {\em arXiv preprint arXiv:1910.10093}, 2019.

\bibitem{zhu2020reconstructing}
Luyang Zhu, Konstantinos Rematas, Brian Curless, Steven~M Seitz, and Ira
  Kemelmacher-Shlizerman.
\newblock Reconstructing nba players.
\newblock In {\em ECCV}, 2020.

\end{thebibliography}
}

\end{document}